\documentclass[10pt,twocolumn,letterpaper]{article}

\usepackage[final]{iccv}      

\definecolor{iccvblue}{rgb}{0.21,0.49,0.74}
\usepackage[pagebackref,breaklinks,colorlinks,allcolors=iccvblue]{hyperref}

\title{EfficientMT: Efficient Temporal Adaptation for Motion Transfer in Text-to-Video Diffusion Models}

\author{Yufei Cai\textsuperscript{\rm 1,\rm 2,\rm 3}, Hu Han\textsuperscript{\rm 1,\rm 2,\rm 3}, Yuxiang Wei\textsuperscript{\rm 4}, Shiguang Shan\textsuperscript{\rm 1,\rm 2,\rm 3}, Xilin Chen\textsuperscript{\rm 1,\rm 2,\rm 3}\\
\textsuperscript{\rm 1}Key Laboratory of Intelligent Information Processing, Chinese Academy of Sciences (CAS)\\
\textsuperscript{\rm 2}Institute of Computing Technology, CAS\\
\textsuperscript{\rm 3}University of the Chinese Academy of Sciences\\
\textsuperscript{\rm 4}Harbin Institute of Technology\\
{\tt\small caiyufei23@mails.ucas.ac.cn, yuxiang.wei.cs@gmail.com, \{hanhu,sgshan,xlchen\}@ict.ac.cn}
}

\begin{document}
\maketitle
\begin{figure*}[t]
    \centering
    \includegraphics[width=1\linewidth]{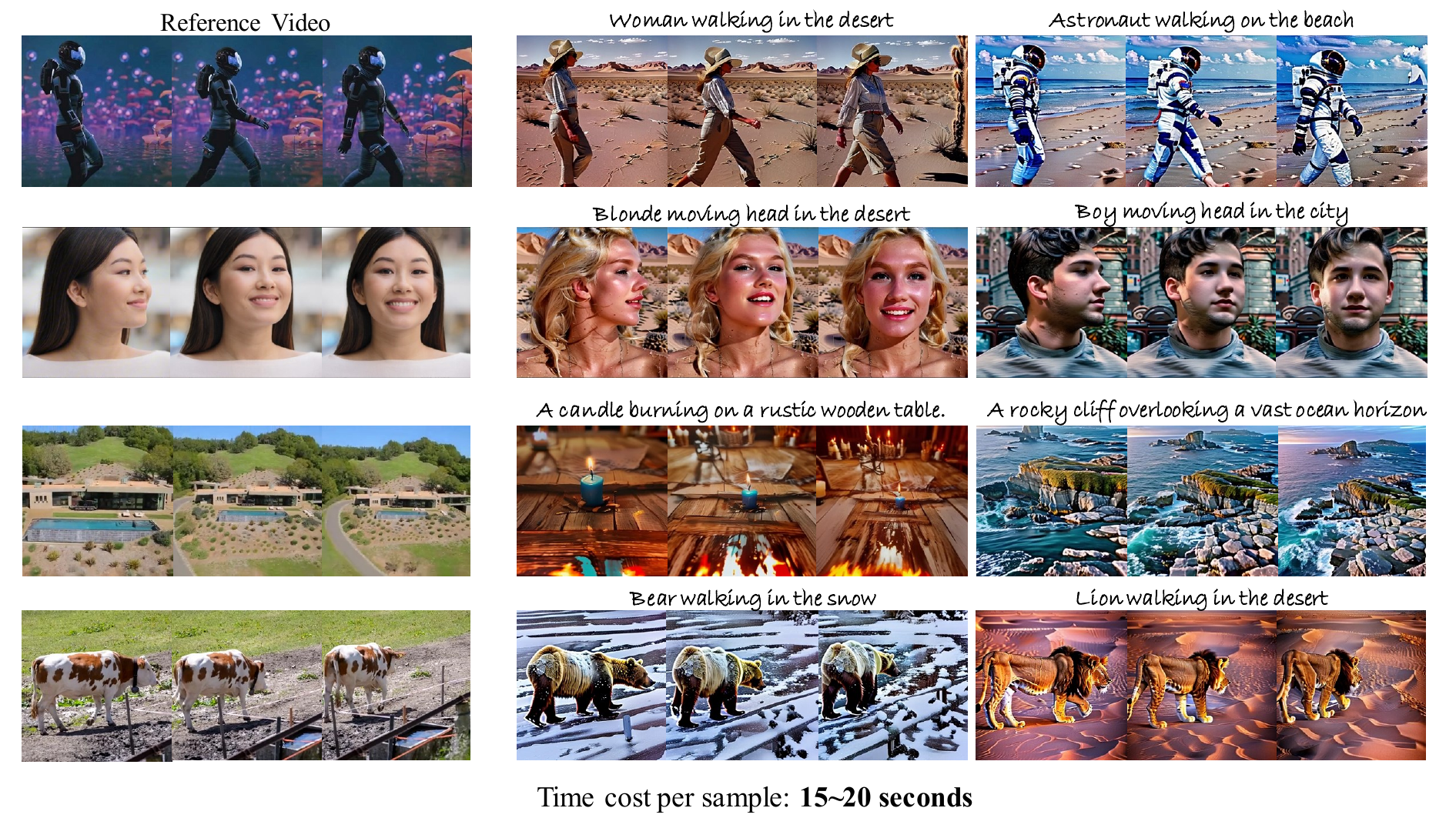}
    \vspace{-1.8em}
    \caption{\textbf{Generation results of our EfficientMT.} Based on a pretrained T2V model, EfficientMT performs zero-shot transfer of both subject and camera motion only in inference time. Please refer to the supplementary materials for better visual evaluation.}
    \vspace{-1.2em}
    \label{fig_head}
\end{figure*}


\begin{abstract}
The progress on generative models has led to significant advances on text-to-video (T2V) generation, yet the motion controllability of generated videos remains limited. 
%
%
Existing motion transfer methods explored the motion representations of reference videos to guide generation.
Nevertheless, these methods typically rely on sample-specific optimization strategy, resulting in high computational burdens.
In this paper, we propose \textbf{EfficientMT}, a novel and efficient end-to-end framework for video motion transfer. 
By leveraging a small set of synthetic paired motion transfer samples, EfficientMT effectively adapts a pretrained T2V model into a general motion transfer framework that can accurately capture and reproduce diverse motion patterns. 
Specifically, we repurpose the backbone of the T2V model to extract temporal information from reference videos, and further propose a scaler module to distill motion-related information.
Subsequently, we introduce a temporal integration mechanism that seamlessly incorporates reference motion features into the video generation process.
After training on our self-collected synthetic paired samples, EfficientMT enables general video motion transfer without requiring test-time optimization.
Extensive experiments demonstrate that our EfficientMT outperforms existing methods in efficiency while maintaining flexible motion controllability.
Our code will be available \url{https://github.com/PrototypeNx/EfficientMT}.

\end{abstract}    

\section{Introduction}
\label{sec:intro}


Recent advancements in generative models~\cite{GAN, DDPM} have significantly advanced visual generation tasks. 
The large-scale pretrained text-to-image (T2I)~\cite{GLIDE, DALLE2, Imagen, LDM} and text-to-video (T2V) models~\cite{animatediff, videocrafter2, latte, alignyourlatent} have demonstrated exceptional capabilities in producing high-quality images and videos, facilitating various downstream applications. 
Building on these models, numerous studies~\cite{controlnet, dreambooth, ipadapter, videocomposer, svd} have focused on improving the controllability of pretrained models by incorporating diverse and user-friendly control conditions into the generation process. 

\begin{figure}[t]
    \centering
    \includegraphics[width=1\linewidth]{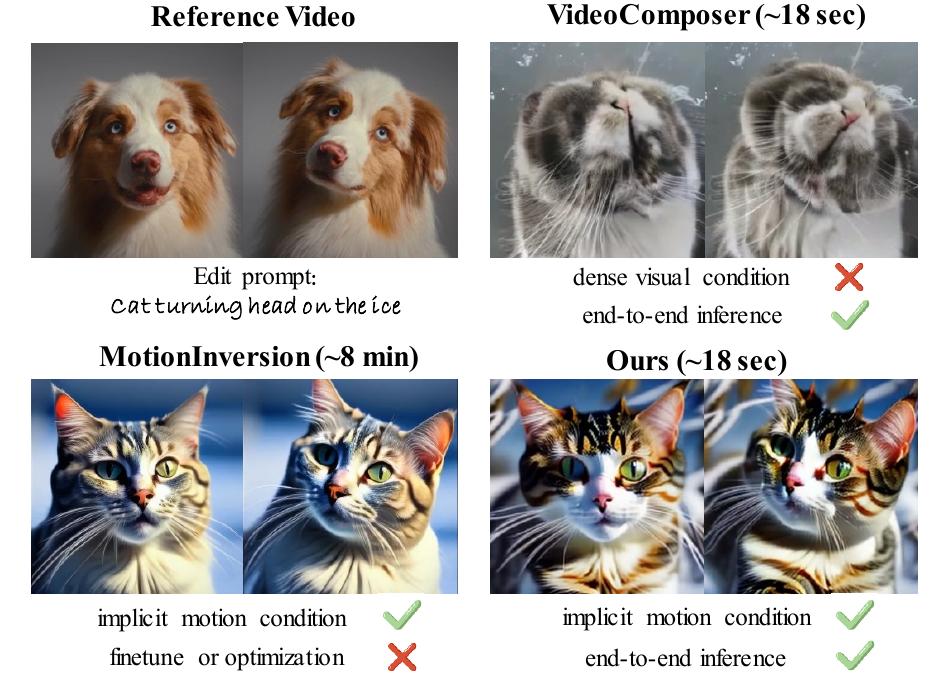}
    \caption{\textbf{Comparison of methods.} Our method inherits strength of both methods, achieving efficient and flexible motion transfer.}
    \label{fig:motivation}
    \vspace{-1.2em}
\end{figure}

Motion is a crucial aspect of temporal content in video generation, yet its precise control remains a significant challenge. 
To address this, numerous efforts have been devoted to video motion transfer, which aims to adapt motion patterns from a reference video to a new subject and scene.
For example, some controllable video generation methods~\cite{videocomposer,controlavideo} introduce dense visual conditions (\eg, depth, sketch, and optical flow) into T2V models.
These models are trained end-to-end to leverage strict frame-by-frame structural guidance for reconstructing motion patterns.
However, dense visual descriptors often impose excessive structural constraints, limiting the model's ability to transfer across different scenes (as shown in Fig.~\ref{fig:motivation}). 
Alternatively, trajectory-based methods~\cite{motionctrl, draganything} offer more intuitive and interactive motion control conditions.
Despite their user-friendly controllability, these methods struggle to achieve fine-grained motion control.

Another line of research~\cite{DMT, VMC, motiondirector,motionclone, MOFT} investigates how motion is encoded within T2V models, aiming to disentangle and extract implicit motion representations from reference videos. 
These representations then serve as motion guidance to refine the generation process, resulting in the desired videos.
Although these methods demonstrate promising motion transfer capabilities, they typically rely on an optimization-based framework. 
Each reference video requires a sample-specific optimization step, significantly increasing computational overhead compared to fully end-to-end generative frameworks~\cite{videocomposer, controlavideo}.

To address the above issues, we propose \textbf{EfficientMT}, a novel and efficient end-to-end framework for video motion transfer. 
By leveraging the prior knowledge of pretrained T2V model, EfficientMT effectively adapts it into a general motion transfer framework using a small set of synthetic paired motion transfer samples.
The adapted model is capable of accurately capturing and reproducing diverse motion patterns while maintaining superior efficiency.
To this end, our EfficientMT first extracts motion information from reference videos, and then incorporates reference motion into the video generation process.
Specifically, we utilize the features from the temporal attention layers of a pretrained T2V model to represent the temporal information of reference videos, which has been demonstrated to contain rich motion details~\cite{motionclone, motioninversion}. 
To disentangle motion-irrelevant information, we further propose the scaler module.
It predicts a fine-grained scale map for the reference features, adaptively filtering out irrelevant information and mitigating its interference with motion synthesis.
Furthermore, we introduce a temporal integration mechanism that seamlessly incorporates reference motion features into the video generation process.
In particular, we concatenate these motion features with the key and value components in the temporal attentions, thereby providing explicit motion information to the model.
During training, both the temporal attention layers and the scaler modules are finetuned simultaneously.


To train EfficientMT, paired video data is required.
However, collecting motion-matched video pairs in real-world scenarios is highly challenging. 
To address this, we utilize baseline methods~\cite{motionclone,motioninversion} to synthesize paired motion transfer data, followed by data cleansing and manual filtering to obtain high-quality motion transfer samples for training.
Experiments demonstrate that our proposed framework can effectively adapt a pretrained T2V model into a general motion transfer model using only a few hundred synthesized motion transfer samples.

Our contributions can be summarized as follows:
\begin{itemize}
    \item We propose EfficientMT, an efficient video motion transfer framework that adapts a pretrained T2V model into an end-to-end motion transfer model using only a small set of synthesized data samples.
    \item To achieve efficient adaptation, we design an effective motion integration strategy. We reuse the backbone for reference feature extraction and introduce temporal integration mechanism combined with a scaler to distill motion guidance into the generation process.
    \item Extensive experiments show that our EfficientMT achieves a faster generation process while exhibiting robust motion transfer capabilities.
\end{itemize}

\section{Related Work}
\label{sec:relatedworks}

\subsection{Text-to-Video Diffusion Models}
Diffusion model \cite{DDPM} is a parametric neural network that approximates complex high-dimensional distributions through an iterative denoising process, making it widely adopted in visual generation. With the advancements of the classifier-free guidance technique \cite{cfg} and language models \cite{clip,t5}, subsequent research has explored text-conditioned visual generation. Leveraging large-scale text-image pairs \cite{LAION-400M}, pretrained text-to-image diffusion models \cite{LDM, DALLE2, Imagen} have demonstrated unprecedented generative performance. Extending this capability to video, AnimateDiff \cite{animatediff} and Align Your Latents\cite{alignyourlatent} integrate temporal attention modules into Stable Diffusion (SD) to inflate it into a text-to-video (T2V) model. Methods like ModelScope \cite{modelscope}, VideoCrafter \cite{videocrafter2}, and VDM \cite{vdm} initialize models with SD priors and retrain them using text-video pairs to achieve video generation.
Recent works such as SVD \cite{svd} and Open-Sora \cite{opensora} introduce novel architectures trained on large-scale text-image and video data through multi-stage training, pushing the boundaries of video generation quality. However, despite their strong generative performance, these T2V models rely solely on text prompts, making it challenging to exert fine-grained control over the generated content. 


\subsection{Video Motion Transfer}
Video motion transfer aims to enable generated content to faithfully reproduce a specified motion pattern. Pioneering works, such as Tune-A-Video \cite{tuneavideo}, inflate pretrained T2I models into personalized video models via sample-specific fine-tuning. Other methods \cite{videocomposer, controlavideo, controlvideo} incorporate dense visual conditions to reconstruct motion patterns. Despite achieving efficient inference, the redundant structural constraint severely limits generalization across diverse subjects.
Trajectory-based methods, such as MotionCtrl \cite{motionctrl}, enable control over coarse-grained motion patterns, eg., camera movements and subject transitions, but struggle with fine-grained action control. Later research employs additional optimization strategies to capture specific motion patterns. DreamVideo \cite{dreamvideo} and MotionDirector \cite{motiondirector} insert LoRA \cite{lora} modules into T2V models to disentangle texture and motion, while VMC \cite{VMC} and MotionInversion \cite{motioninversion} fine-tune the temporal module to distill motion features. DMT \cite{DMT}, MotionClone \cite{motionclone}, and MOFT \cite{MOFT} employ training-free strategies, optimizing intermediate noise maps during denoising using motion guidance. These approaches yield more robust motion transfer, but the additional optimization step significantly increases computational overhead.
In this work, we integrate the strengths of these methods by adapting a pretrained T2V model into an end-to-end motion transfer framework. Our approach leverages implicit motion representations to enable robust motion transfer, while simultaneously ensuring fast end-to-end generation.

\begin{figure*}[t]
    \centering
    \includegraphics[width=1\linewidth]{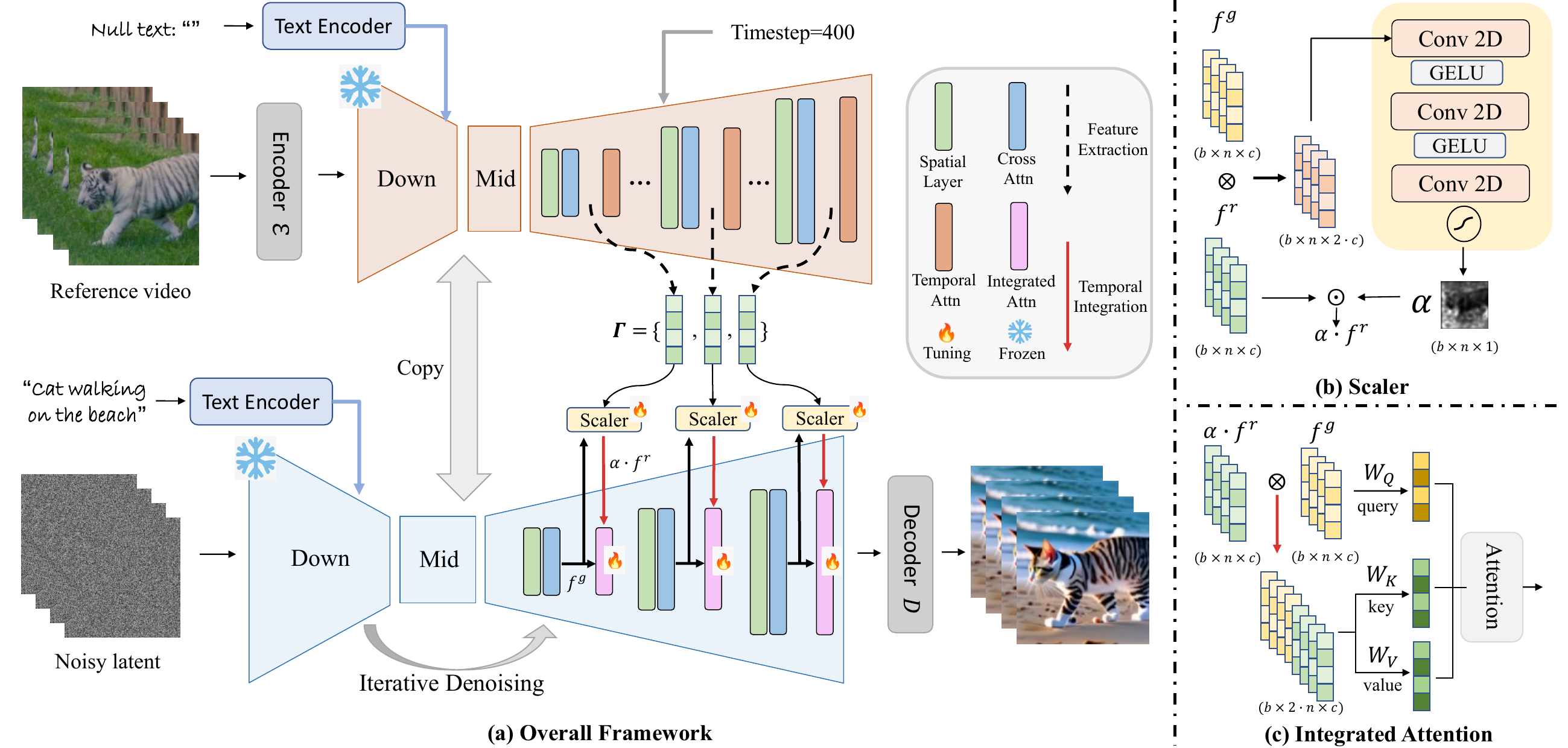}
    \vspace{-1.8em}
    \caption{\textbf{Overview of our EfficientMT.} \textbf{(a)}: We reuse the backbone of the T2V model to extract reference features, which are then injected into the temporal attention layers of the upsampling stage through a temporal integration mechanism. \textbf{(b)}: The scaler predicts a fine-grained scale map for the reference features, filtering out irrelevant information. \textbf{(c)}: The temporal integration concatenates the features along the temporal axis, while the query is projected from the origin, the key and value are obtained from the integrated features.}
    \vspace{-1.2em}
    \label{fig_pipeline}
\end{figure*}


\section{Proposed Method}

Given a reference video $x$ with certain motion $M$ and a text prompt $y$, our goal is to generate a new video $x'$ that aligns with the description in $y$ while faithfully replicating the motion pattern $M$ observed in $x$. In the following, we first present an overview of the T2V models used in our approach (Sec 3.1). Next, we introduce the motion feature extraction method (Sec 3.2) and explain our temporal integration mechanism (Sec 3.3). Finally, we describe the construction for paired motion transfer data (Sec 3.4).

\subsection{Preliminary}
We employ AnimateDiff \cite{animatediff} and VideoCrafter2 \cite{videocrafter2} as our T2V base models. These models extend the structure of T2I models to accommodate video generation. Initially, an auto-encoder ($\mathcal{E}(\cdot)$,$\mathcal{D}(\cdot)$) is trained to map videos into a low-dimensional latent space. $\mathcal{E}$ encodes the video $x$ to the latent space $z=\mathcal{E}(x)$ and the decoder 
$\mathcal{D}$ reconstructs the latent variables back into the natural video space $\mathcal{D}(\mathcal{E}(x)) \approx x$. A denoising network $\epsilon_\theta$ is trained in the latent space to iteratively predict the latent noise based on the text condition $y$. The model is trained using mean squared error loss \cite{DDPM} to optimize the diffusion process:
\begin{equation}
\small
\mathcal{L}_\text{MSE} = \mathbb{E}_{z\sim\mathcal{E}(x), y, \epsilon \sim \mathcal{N}(0, 1), t }\Big[ \Vert \epsilon - \epsilon_\theta(z_{t},t, \tau_\theta(y)) \Vert_{2}^{2}\Big] \, ,
\label{eqn:diffusion}
\end{equation}
where $\epsilon$ denotes the unscaled noise, $t$ is the time step, $z_t$ is the latent noised to time $t$, and $\tau_\theta(\cdot)$ represents the pretrained text encoder.

The temporal attention layer is inserted into the T2V model to enable temporal information exchange. Specifically, the temporal axis of the latent feature $f$ is first reshaped and moved to the front, treating each frame as a token. A projection layer then transforms these tokens to obtain the query $Q=W_Q\cdot f$, key $K=W_K\cdot f$, and value $V=W_V\cdot f$. Then, the attention mechanism \cite{attention} is conducted along the temporal dimension by:
\begin{equation}
\text{Attention}(Q,K,V) = \text{Softmax}(\frac{QK^T}{\sqrt{d'}})V ,
\label{eqn:attn}
\end{equation}
where $d'$ is the output dimension of key and query features.

\subsection{Motion Representation}
As shown in Figure \ref{fig:motivation}, optimization-based methods \cite{motioninversion, motionclone} enable better transfer capability, as the distilled implicit motion representations are inherently more robust. While other methods \cite{videocomposer,controlavideo} that incorporate dense visual conditions can easily collect large amounts of paired annotated data for end-to-end training, which is particularly appealing in terms of inference efficiency.

Based on this, we hypothesize that it is possible to train an end-to-end transfer framework using implicit motion representations with a small set of paired motion transfer samples. This framework should efficiently capture specific motion patterns in a zero-shot manner during the generation process. It becomes crucial to maximally leverage the priors from the pretrained model to ensure efficient training. As the extracted reference features should be well-aligned with the generation process, we opt to directly reuse the pretrained T2V backbone as a feature extractor $\hat{\epsilon}$. Prior work \cite{motiondirector} has demonstrated  that the spatial and temporal modules exhibit distinct functional roles, so we extract the inputs of the temporal attention layer as the reference features $\mathit{\Gamma}$, which will be injected at the corresponding locations in the generation branch:
\begin{equation}
\mathit{\Gamma} = \hat{\epsilon}(x,t,\tau_\theta(y^r))=\{f\}_{upblocks}^{temporal},
\label{eqn:ref_feat}
\end{equation}
We set the timestep $t$ to the latter stages of the denoising process and send the reference prompt $y^r$ as \textit{null-text} to better capture motion dynamics.

\begin{figure}[t]
    \centering
    \includegraphics[width=1\linewidth]{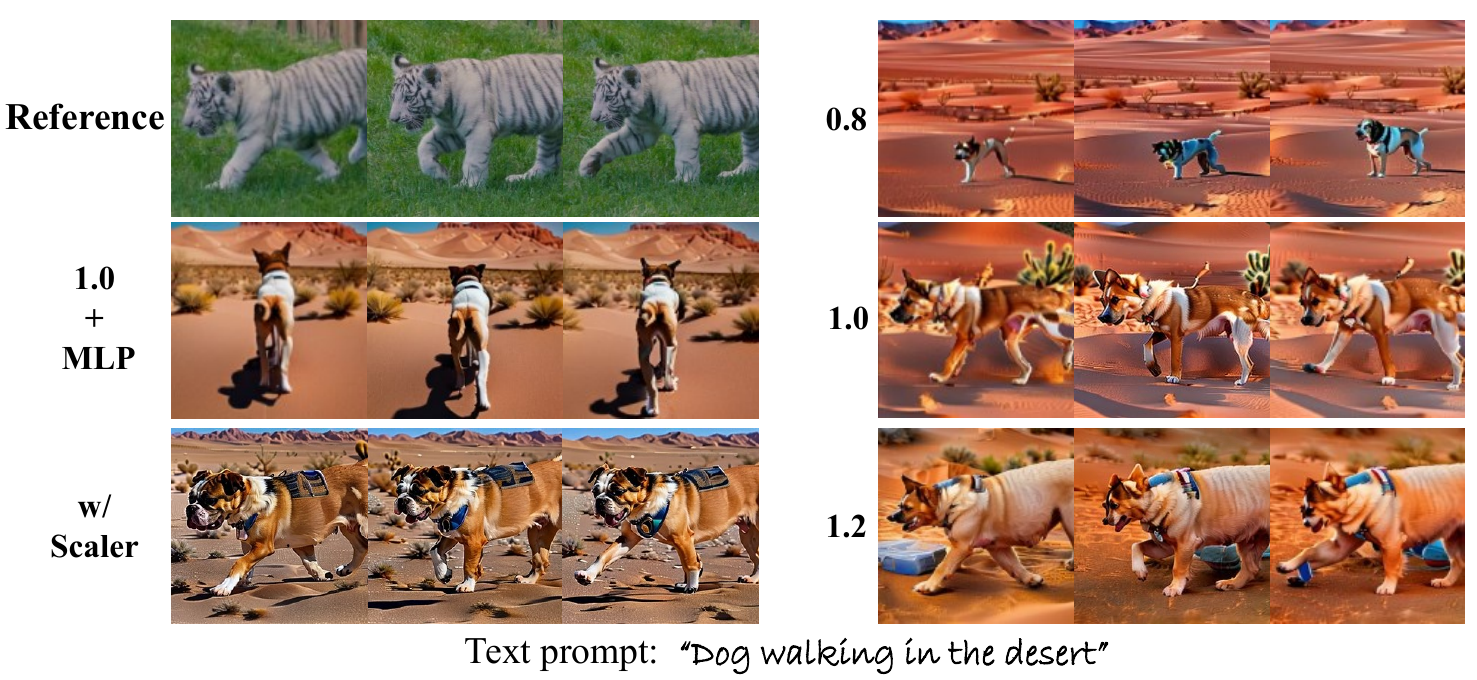}
    \caption{\textbf{Visual comparisons on the effect of the integration scale}. As the injection scale of reference features increases, the control over the generated content becomes more pronounced. Introducing a scaler enhances the robustness of the generation.}
    \label{fig:scale}
    \vspace{-1.2em}
\end{figure}

Prior works \cite{MOFT, DMT} observed that motion-related information is inherently encoded within the intermediate features of T2V models. However, empirical analysis reveals that manually disentangling motion representations from these features remains challenging.
A natural approach is to apply simple transformations to the extracted features, enabling adaptive refinement of motion-relevant components. However, experimental results indicate that transformations introduce a significant domain gap, severely increasing training difficulty and disrupting motion fidelity (see Figure \ref{fig:scale}). Furthermore, we observe that adjusting the injection scale of features can effectively influence the motion preservation. Inspired by \cite{smartcontrol}, we introduce a fine-grained scale predictor for the reference features. This scaler $\mathcal{S}(\cdot)$ estimates a scale map $\alpha$, enhancing motion representation while suppressing irrelevant information.
Specifically, for the reference features $f^{r}\in\mathit{\Gamma}$ and the corresponding features $f^{g}$ in the generation branch, we concatenate them along the channel axis and feed it into the predictor $\mathcal{S}$:
\begin{equation}
\alpha = \mathcal{S} (f^{r}\otimes f^{g}) \in \mathbb{R}^{h\times w\times n \times 1},
\label{eqn:scale}
\end{equation}
where $f^{r},f^{g}\in \mathbb{R}^{h\times w\times n \times c}$, $\otimes$ denotes concatenation, and $n$ is the number of video frames. Then we apply the scale map $\alpha$ to the reference feature $f^{r}$ by $\hat{f}^r=\alpha\cdot f^{r}$, filtering irrelevant information adaptively.

\subsection{Temporal Integration}
The features from two branches are integrated within the temporal attention layers. We concatenate the filtered reference feature $\hat{f}^r$ with $f^{g}$ along the temporal axis:
\begin{equation}
f^{int} = f^{g} \otimes \hat{f}^{r}  \in \mathbb{R}^{h\times w\times 2\cdot n \times c},
\label{eqn:concat}
\end{equation}
The query is projected from the original $f^{g}$, while we obtain the key and value from integrated feature $f^{int}$:
\begin{equation}
Q = W_Q\cdot f^{g}, K = W_K\cdot f^{int}, V = W_V\cdot f^{int},
\label{eqn:qkv}
\end{equation}
This operation seamlessly incorporates reference features with inherited structure. It treats the reference feature as additional video frames, providing auxiliary guidance for generation. When temporal attention layers perform \ref{eqn:attn} by replacing the above terms, the training objective encourages the query to aggregate motion information from the reference frames, facilitating the expected transfer reconstruction.

During training, we only finetune the integrated temporal attention layers. In contrast to previous conclusions \cite{MOFT, motionclone}, we found that injecting features into only one upsampling block is insufficient to capture general motion patterns. Efficient zero-shot transfer capabilities require maintenance of more reference features. Therefore, we integrate reference features from all upsampling blocks to better accommodate motion pattern extraction. Specifically, we modify the projection matrices $W_Q$, $W_K$, and $W_V$ of the query, key, and value in all temporal attention layers of the upsampling stage. We employ the standard diffusion loss as the training objective:
\begin{equation}
\small
\mathcal{L} = \mathbb{E}_{z\sim\mathcal{E}(x'), y, \epsilon \sim \mathcal{N}(0, 1), t, \mathit{\Gamma}\sim\hat\epsilon(x)}\Big[ \Vert \epsilon - \epsilon_\theta(z_{t},t, \tau_\theta(y),\mathit{\Gamma}) \Vert_{2}^{2}\Big] \, ,
\label{eqn:my_diffusion}
\end{equation}

\subsection{Motion Transfer Data Construction}
The framework design outlined requires paired motion transfer data to guide the model in capturing motion patterns. However, collecting video pairs with consistent motion patterns across scenes is extremely challenging in the real world. We leverage existing advanced motion transfer methods to construct data pairs to address this. Specifically, we consider adopting MotionClone \cite{motionclone} and MotionInversion \cite{motioninversion}, both of which employ sample-specific optimization strategies.
We collect real reference videos from the RealEstate10K \cite{RealEstate10K}, DAVIS datasets \cite{davis} and website each exhibiting a specific motion pattern. These videos are then paired with textual descriptions generated by BLIP \cite{blip}, formatted as \texttt{"subject, action, background"} for subject. Subsequently, we utilize ChatGPT to produce multiple edited versions mainly altering the \texttt{"subject"} and \texttt{"background"} for subject and rewrite the descriptions for camera motion. These edited descriptions are used with baseline methods to generate motion transfer samples.
To ensure that the generated samples exhibit a well-aligned motion pattern and temporal continuity, we filter them based on two criteria: motion alignment and temporal consistency. Specifically, we use the motion fidelity score \cite{DMT} to measure motion alignment, and CLIP's \cite{clip} image inter-frame features for assessing temporal consistency. After that, we conducted a manual selection process, ultimately obtaining over 150 high-quality samples based on the two pretrained T2V models respectively.

\begin{figure*}[t]
    \centering
    \includegraphics[width=1\linewidth]{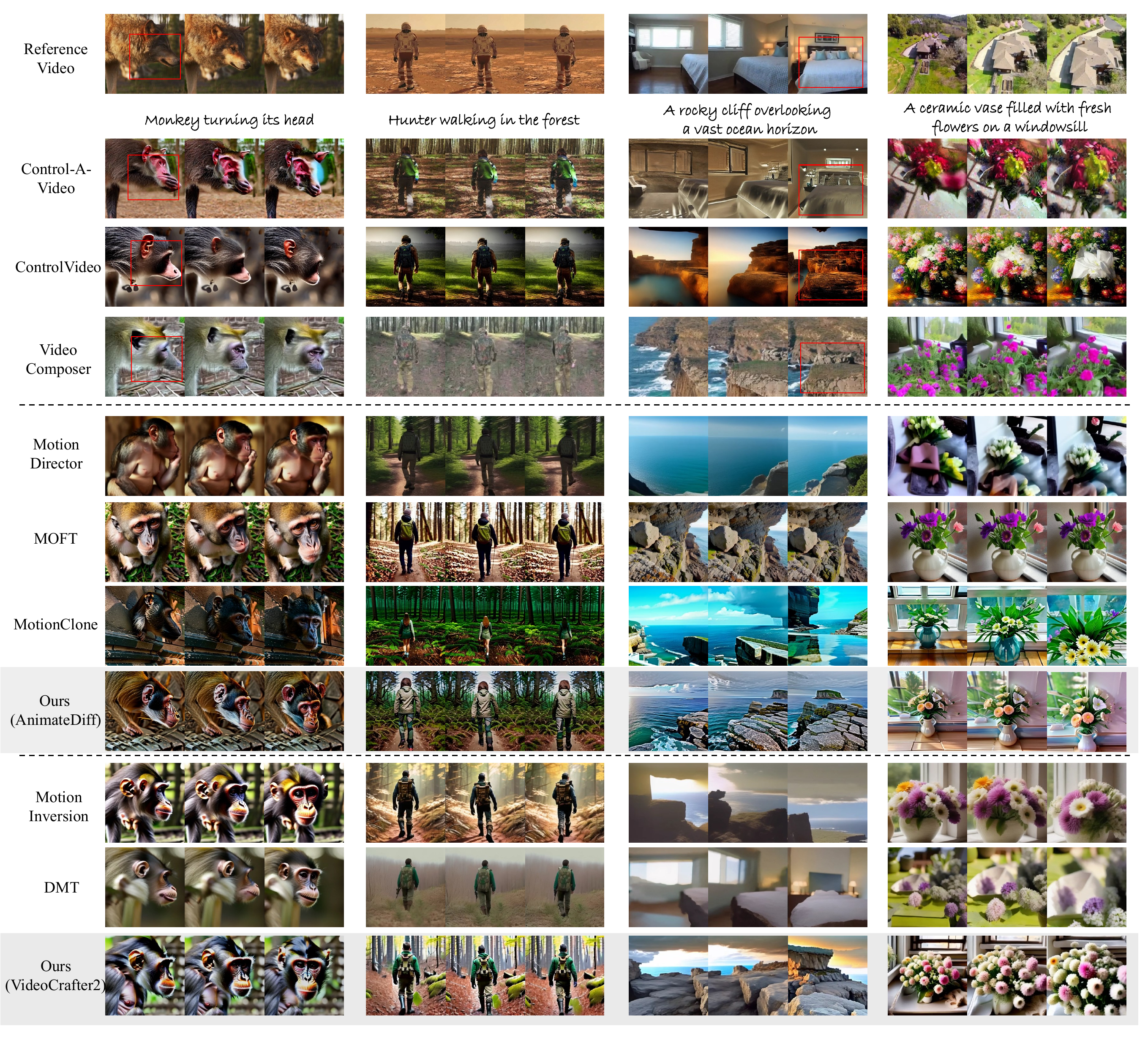}
    \vspace{-1.8em}
    \caption{\textbf{Visual comparisons.} Our EfficientMT enables the unified transfer of subject and camera motion. Compared to state-of-the-art methods, our method offers superior editing flexibility and motion fidelity. Zoom up for a better view.}
    \vspace{-1.2em}
    \label{fig_quality}
\end{figure*}

\section{Experiments}
\label{sec:experiments}

\subsection{Experimental Settings}
\textbf{Datasets.} We collect a total of 217 real motion reference videos from the RealEstate10K \cite{RealEstate10K}, DAVIS dataset \cite{davis} and website, covering over 40 different categories of subjects and over 30 different camera motions. During the training data construction, we generated 5$\sim$10 editing prompts for each subject video in the training set, resulting in over 1200 constructed videos. After filtering and manual selection, we obtain 150 high-quality samples of 31 reference videos in the training set. For testing, we similarly generated 5$\sim$10 editing prompts for 82 videos in the test set, yielding more than 400 test samples.

\noindent \textbf{Baselines.}
We compare our method with both dense-visual-condition-based methods: VideoComposer \cite{videocomposer}, Control-A-Video \cite{controlavideo}, ControlVideo \cite{controlvideo}, and optimization-based methods: MotionDirector \cite{motiondirector}, MotionClone \cite{motionclone}, DMT \cite{DMT}, MotionInversion \cite{motioninversion}, and MOFT \cite{MOFT}. The open-source code for these methods is implemented based on one of the two T2V base architectures \cite{animatediff, videocrafter2}. To ensure fairness, we adapt our EfficientMT to both architectures and compare it with the corresponding methods. We also incorporate a realistic visual style LoRA module for AnimateDiff. Please refer to the \textit{suppl.} for more implementation details.

\noindent \textbf{Metrics.} We evaluate motion transfer methods from four aspects:
(i) \textit{Text Alignment}: We use CLIP to extract visual features frame by frame and calculate the average cosine similarity with the corresponding text features.
(ii) \textit{Temporal Consistency}: This is measured by calculating the average cosine similarity of the CLIP visual features between every two frames in the video.
(iii) \textit{Motion Fidelity}: Following DMT \cite{DMT}, we apply the co-tracking method \cite{cotracker} to obtain point trajectories that match across videos. We measure the correspondence between these point trajectories.
(iv) \textit{Time Cost}: This measures the total time required to capture a new reference motion pattern. It is reasonable to include the additional fine-tuning or optimization steps in count for optimization-based methods.
\begin{table*}[t]
    \centering
    \setlength{\tabcolsep}{3mm}
    \resizebox{0.8\linewidth}{!}{
        \begin{tabular}{ccccc}
            \toprule
            Method & Temporal Consistency $(\uparrow)$ & Text Alignment $(\uparrow)$ & Motion Fidelity $(\uparrow)$ & Time Cost $(\downarrow)$\\
            \midrule
            ControlVideo \cite{controlvideo} & 0.9213& 0.2483 & 0.5533  & 80s 
            \\
            VideoComposer \cite{videocomposer} & 0.9192& 0.2635 & 0.6356  & 18s 
            \\
            Control-A-Video \cite{controlavideo} & 0.9172& 0.2438 & 0.6225  & 25s 
            \\
            \cline{1-5}
            MotionDirector \cite{motiondirector} & \textbf{0.9327}& 0.2525 & 0.8361  & 473s 
            \\ 
            MotionClone \cite{motionclone} & 0.9108 & \underline{0.2637} & \textbf{0.8569} & 190s
            \\ 
            MOFT \cite{MOFT} & 0.9283 & 0.2581 & 0.7698 & \underline{127s}
            \\
            Ours (AnimateDiff) & \underline{0.9291} & \textbf{0.2712} & \underline{0.8470} & \textbf{16s}
            \\
            \cline{1-5}
            DMT \cite{DMT} & 0.9275 & 0.2479 & 0.6974   & \underline{203s}
            \\
            MotionInversion \cite{motioninversion} & \underline{0.9329} & \underline{0.2558} & \textbf{0.7373}  & 418s
            \\
            Ours (VideoCrafter2) & \textbf{0.9456} & \textbf{0.2677} & \underline{0.7116}  & \textbf{21s}
            \\
            \bottomrule
        \end{tabular}
    }
    \caption{\textbf{Quantitative comparisons.} Our method achieves competitive video generation quality while offering a significant advantage in computational efficiency.}
    \label{tab:quantitative}  
\end{table*}

\begin{table}[t]
    \small
    \centering
    \setlength{\tabcolsep}{3mm}
    \begin{tabular}{ccccc}
        \toprule
        Method & VC & MD & MC & MI\\
        \midrule
        Temporal Consistency & 75.3 & 48.4 & 55.9  & 57.0 \\ 
        Text Alignment & 80.6 & 64.5 & 57.0 & 61.3  \\ 
        Motion Fidelity & 75.3 & 78.4 & 49.5 & 58.1 \\
        \bottomrule
    \end{tabular}
    \caption{\textbf{User study.} The numbers indicate the percentage (\%) of volunteers who favor the results of our method over those of the competing methods based on the given question.}
    \label{tab5:user_study}  
    \vspace{-1.2em}
\end{table}

\subsection{Qualitative Evaluation}
First, we qualitatively compare EfficientMT with existing methods. The comparison results are illustrated in Figure \ref{fig_quality}. Compared to methods based on dense visual conditions (the first three rows), our EfficientMT eliminates the rigid structural constraints, enabling more flexible transfer (\eg, the contours of the monkey in the first column). Our method also supports the transfer of camera motion, a task that is challenging for the above methods (\eg, the content within the red boxes in the first and fourth columns). This indicates that our method captures representations that are faithfully related to motion patterns rather than simple shape correspondences. Compared to optimization-based methods, our approach significantly outperforms methods such as MOFT and DMT in terms of motion fidelity. Furthermore, our qualitative results are also highly competitive compared to other optimization-based methods requiring longer optimizing times. Comparison with trajectory-based method \cite{motionctrl} and more qualitative results can be found in the \textit{suppl}.

\subsection{Quantitative Evaluation}
We further conducted a quantitative evaluation. The quantitative results are shown in Table \ref{tab:quantitative}. Although the methods based on the dense visual condition (the first three rows) have a similar time cost to our method, they are significantly inferior to ours in terms of quality metrics, which aligns with the conclusions drawn from the qualitative experiments. Compared to optimization-based methods, our approach demonstrates a substantial improvement in time efficiency to capture new motion patterns. Moreover, our method achieves the best text alignment, enabling more flexible transfer. Our approach also exhibits exceptional performance in terms of motion fidelity and temporal consistency. Therefore, EfficientMT strikes a balance between time efficiency and generation quality, making it more suitable for practical applications in real-world scenarios.

\begin{figure}[t]
    \centering
    \includegraphics[width=1\linewidth]{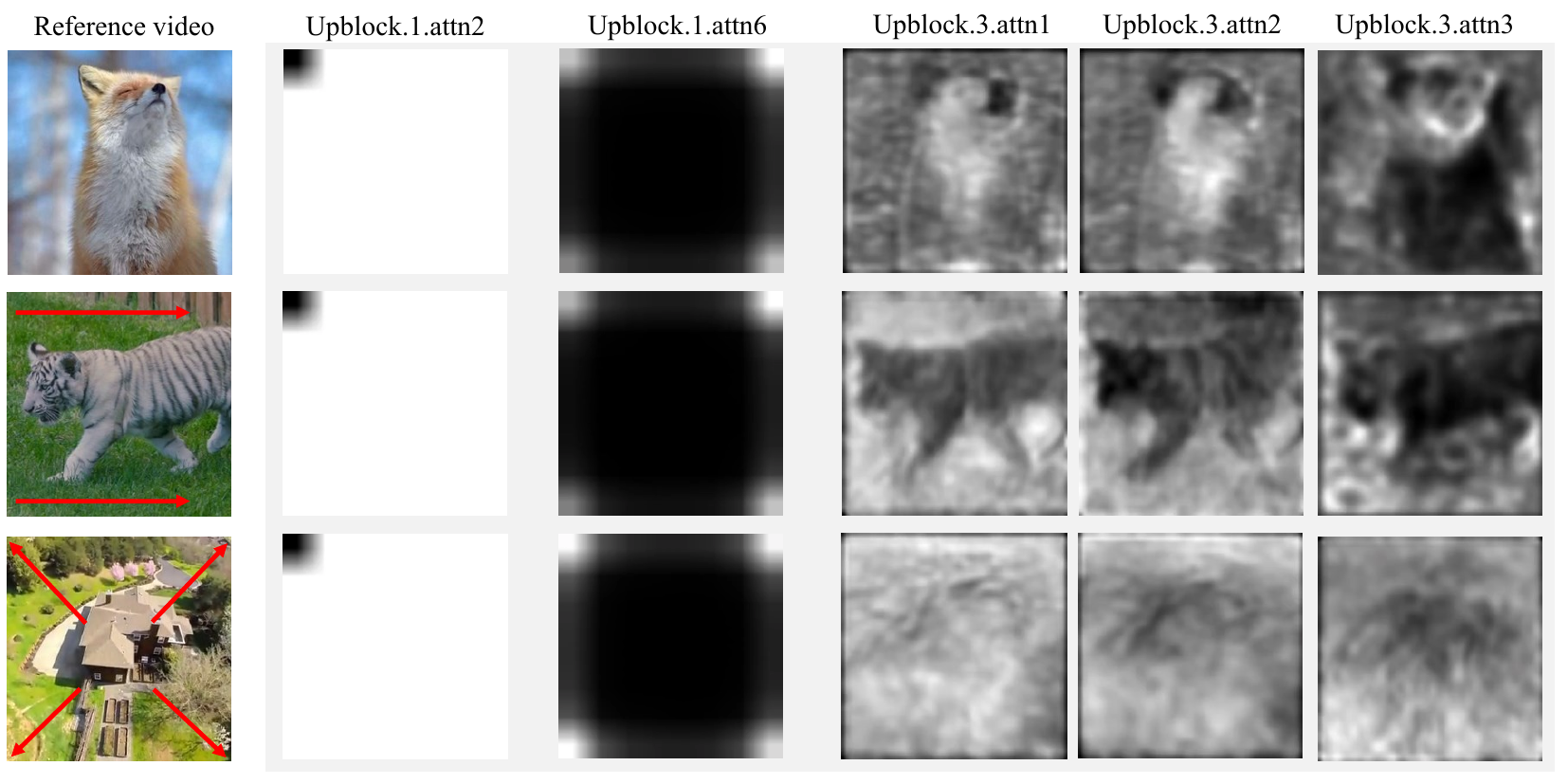}
    \caption{\textbf{Visualization of the scale map.} The scaler adaptively learns the feature injection strength, exhibiting a certain pattern across different references.}
    \label{fig:scalemap}
    \vspace{-1.2em}
\end{figure}
\noindent \textbf{User Study.}
We performed a user study to compare our approach with existing methods.
Given a reference video, users were presented with two synthesized video, asked to select the better one from three aspects:
i) Temporal Consistency: ``Which video is more consistent across frames?'';
ii) Text alignment: ``Which video is more consistent with the text?'';
iii) Motion Fidelity: ``Which video better reproduces the motion pattern in the reference?''.
For each evaluated view, we employ 31 users, each user is asked to answer 36 randomly selected questions.
As shown in Table~\ref{tab5:user_study}, our method receives more preference than other methods.

\begin{figure*}[t]
    \centering
    \includegraphics[width=1\linewidth]{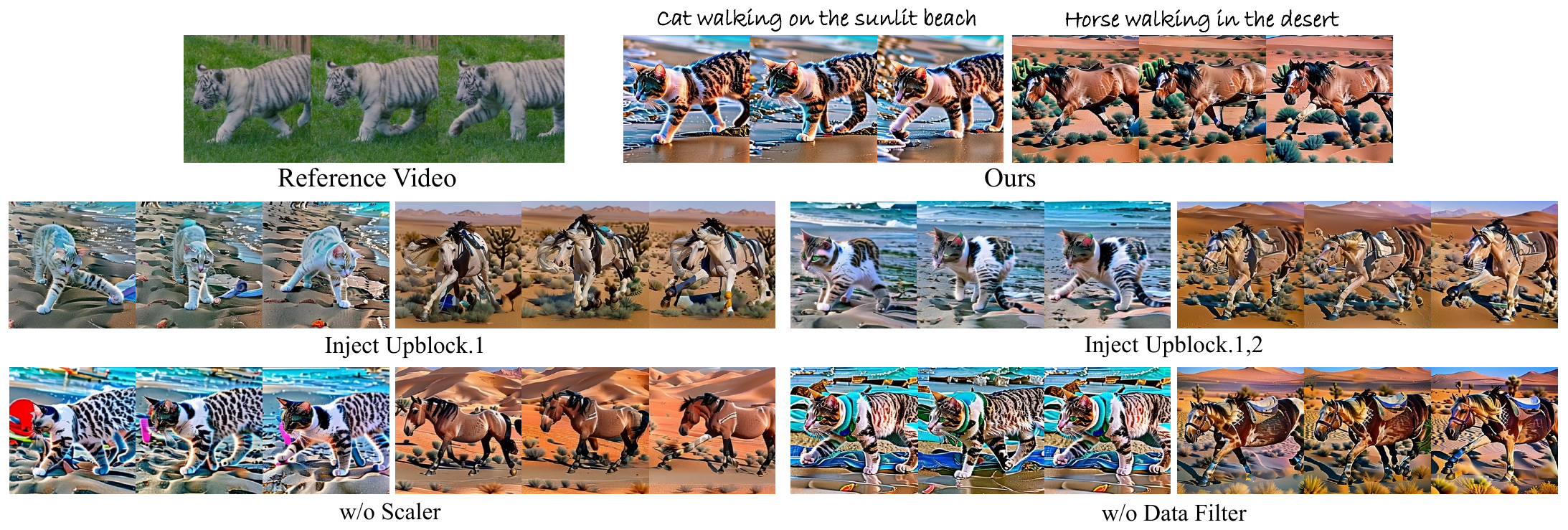}
    \vspace{-1.8em}
    \caption{\textbf{Visualization of the ablating components.}}
    \vspace{-1.2em}
    \label{fig_ablation}
\end{figure*}

\subsection{Ablation Study}
We conducted ablation experiments to analyze the effectiveness of each component, including the scale map, feature injection blocks, and the quality of training data. Figure \ref{fig_ablation} presents the qualitative results after ablating each component, and Table \ref{tab4:ablation} shows the impact of different components on the quantitative metrics. A detailed analysis is provided in the following subsections.

\noindent\textbf{Effect of Scale Map.}
We first removed the scaler and trained the model with direct temporal integration, which is equivalent to setting the scale to a constant value of 1. During the inference, manually adjusting the overall injection ratio significantly impacted the motion fidelity of the generated results. As shown in Figure \ref{fig:scale}, when the scale was reduced to below 1, the results failed to replicate the reference motion pattern, while increasing the scale caused overfitting of the reference subject’s shape and texture, leading to degraded transfer outcomes. This suggests that the reference features contain redundant information, negatively impacting the transfer's generalization. Thus we introduce a scaler to adaptively adjust the injection scale of features. As shown in Figure \ref{fig_ablation}, removing the scaler leads to the emergence of artifacts in the generated results, and motion consistency deteriorates, with a decline in all metrics in the quantitative evaluation in Table \ref{tab4:ablation}.
We also visualize the scale maps of different layers in Figure \ref{fig:scalemap}. The scaler tends to select or discard reference features at specific layers in the low-resolution blocks, with the scale predictions being close to 1 or 0. In the high-resolution blocks, the scaler selectively focuses on the dynamic motion-related region (\eg. moving head, walking legs, and dynamic background) while suppressing the static texture.
This demonstrates that the scale map effectively filters out motion-irrelevant information, improving the robustness of the transfer.

\begin{table}[t]
    \small
    \centering
    \begin{tabular}{cccc}
        \toprule
        Method & TC $(\uparrow)$ & TA $(\uparrow)$ & MF $(\uparrow)$\\
        \midrule
        w/o Scaler & 0.9244 & 0.2638 & 0.8135 \\ 
        w/o Data Filter & 0.9237 & 0.2649 & 0.8278 \\
        Inject Upblock.1 & 0.9013 & \textbf{0.2789} & 0.6374 \\
        Inject Upblock.1,2 & 0.9198 & 0.2755 & 0.7236 \\
        Ours & \textbf{0.9291} & 0.2712 & \textbf{0.8470}\\
        \bottomrule
    \end{tabular}
    \caption{\textbf{Ablation study.} Metrics are represented using abbreviations of their initials.}
    \label{tab4:ablation}  
    \vspace{-1.8em}
\end{table}
\noindent\textbf{Effect of Inject Layer.}
We further investigated the effect of the feature injection blocks. We chose to inject features only into the upblock1, or simultaneously apply injection to both upblock1 and upblock2. As shown in Figure \ref{fig_ablation}, injection into only upblock1 barely replicates the motion, and the generated subject collapses. Moreover, injection into both upblock1 and upblock2 improves motion consistency, but still fails to reach an effective level. From a quantitative perspective (Table \ref{tab4:ablation}), partially injecting features leads to an improvement in text alignment, as fine-tuning a smaller number of model parameters mitigates language drift. However, this approach significantly reduces motion fidelity, which is unacceptable in motion transfer tasks. Therefore, performing temporal integration and fine-tuning all upsampling blocks is both reasonable and necessary.

\noindent\textbf{Effect of Training Data Quality.}
Finally, we verified the necessity of data cleaning during the data construction phase. Failure cases are often generated with the baseline methods, which misguide motion pattern capture. As shown in Figure \ref{fig_ablation},  although the dataset size increases when training with the entire constructed dataset, the generated results are prone to artifacts. It also brings a degradation to the visual quality of the generated frames (\eg, saturation, lighting). The quantitative evaluation shows that the absence of data filtering leads to a decline in various metrics. Therefore, compared to simply increasing the training data size, data quality is more meaningful for improving the model’s performance. More detailed evaluation of the training dataset scale is provided in the \textit{suppl.}

\subsection{Limitations}
\begin{figure}[t]
    \centering
    \includegraphics[width=1\linewidth]{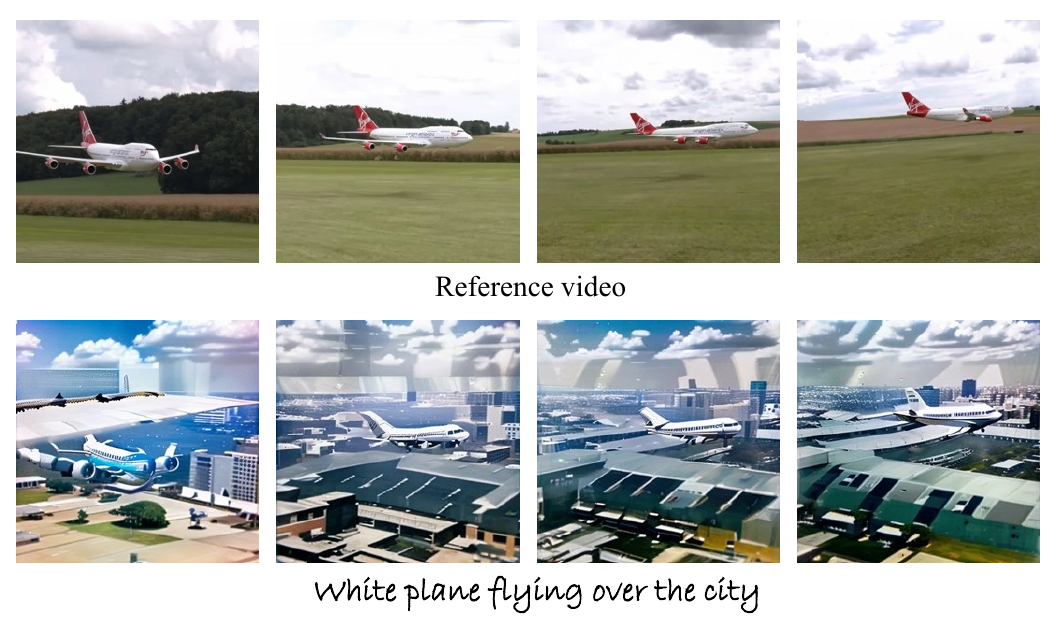}
    \caption{\textbf{Limitations}. Our method struggles to maintain temporal consistency in violently changing motion patterns.}
    \label{fig:limitation}    \vspace{-1.2em}
\end{figure}

As shown in Figure \ref{fig:limitation}, similar to base T2V models, our method struggles to maintain consistency under drastic changing motion patterns, leading to tearing artifacts.

\section{Conclusion}
\label{sec:conclusion}
In this paper, we propose an efficient end-to-end motion transfer framework, named EfficientMT, which adapts a pretrained T2V model into a motion transfer framework with a few synthetic data. To achieve efficient training, we reuse the T2V model's backbone as a feature extractor, introduce scaler to filter motion features, and implement a temporal integration mechanism for information injection. Qualitative and quantitative experiments demonstrate that, compared to existing methods, our method ensures robust motion transfer while achieving faster motion pattern capture. Therefore, EfficientMT is an efficient and versatile motion transfer tool across various T2V models. In the future, we will explore more stable motion transfer techniques under scenarios of violently motion changes.

{
    \small
    \bibliographystyle{ieeenat_fullname}
    \bibliography{main}
}

\end{document}